\documentclass[letterpaper, 10 pt, conference]{ieeeconf}  

\IEEEoverridecommandlockouts                              

\usepackage{graphics} 
\usepackage{epsfig} 
\usepackage{amsmath} 
\usepackage{amssymb}  
\usepackage{xcolor}

\usepackage{subcaption}
\usepackage{cite}
\usepackage{algorithmic}
\usepackage{graphicx}
\newtheorem{theorem}{Theorem}

\usepackage{siunitx}
\usepackage{caption}
\usepackage{algorithm}
\usepackage{algorithmic}
\usepackage{pifont}
\usepackage[table]{xcolor}  
\definecolor{highlightgreen}{RGB}{200,230,201}
\newtheorem{proposition}{Proposition}

\title{\LARGE \bf Selective Sliding Mode Control Enhances Robustness of Aerial Robots Under Disturbances}

\author{Anoop Kiran, Thales C. Silva, and Nora Ayanian
\thanks{All authors are with the School of Engineering and Department of Computer Science, Brown University, Providence, RI 02912, USA. Email: {\tt\footnotesize \{anoop\_kiran, thales\_silva, nora\_ayanian\}@brown.edu}}}

\begin{document}

\maketitle
\thispagestyle{empty}
\pagestyle{empty}

\begin{abstract}

Quadrotor control under disturbances such as wind gusts and payload variations remains challenging. First-order sliding mode control (SMC) and matched/unmatched uncertainty distinction are classical approaches; however, neither tells a practitioner which loops in a cascaded controller should be made robust to sustain disturbances, how to tune those loops, nor what the additional costs are on an embedded processor. Applying robust control uniformly across all axes addresses disturbances indiscriminately, adding tuning effort on axes where no disturbance acts. On inherently underactuated vehicles, disturbances along some axes cannot be opposed directly at all.
We propose a principled methodology for deciding where robust control belongs in a cascaded architecture, with all control running onboard the vehicle. We recover the acting disturbance from PID hover flights under disturbance by inverting the equations of motion, use the recovered signal to identify and quantify the excited axes, and deploy robust control where it counts. Experiments on a Crazyflie 2.1 quadrotor under slung-payload and oblique-wind disturbances show tracking improvements concentrated on the axes the rule selects, and gain sweeps confirm that the switching gain reduces error on matched axes but has no effect on an excited unmatched axis, as the analysis predicts.
\vspace{-1mm}
\end{abstract}


\section{Introduction}

Quadrotors are increasingly deployed for infrastructure inspection, search and rescue, and monitoring, where external disturbances routinely degrade tracking performance and transient stability~\cite{SimonFlowDrone, LeeGeomControl, KiranInfluenceDownwash}. Counteracting these disturbances is complicated by underactuation: four control inputs must govern six degrees of freedom, and quadrotors produce no lateral force directly and must tilt to move sideways. This same structure, however, offers an opportunity. The distinction between disturbances that actuators can directly oppose (\emph{matched}) and those they cannot (\emph{unmatched})~\cite{SinhaMatchedUnmatched, WuL1GeomTracking, CheeUncertainty} offers a natural criterion for where robust control is worth deploying. Yet this structure is rarely exploited, with robust switching laws applied uniformly across all axes rather than only where a disturbance can be opposed~\cite{YazdanshenasQuaternionSMC, GomieroSMCCascade, MeneboNeuroFuzzyASMC}. 

Cascaded PID remains the practical default for its low computational overhead and intuitive tuning~\cite{bitcrazeControllersCrazyflie}. However, its disturbance rejection is limited by the bandwidth-noise tradeoff of fixed-gain linear feedback~\cite{AstromFeedbackSystems}. Integral action removes constant steady-state error asymptotically, so mid-flight payload addition or a sudden crosswind induces a transient excursion before the integrator converges. Raising PID gains for faster rejection amplifies measurement noise and shrinks the stability margin, trading one limit for another. 

\begin{figure}
    \centering    \includegraphics[width=1\linewidth]{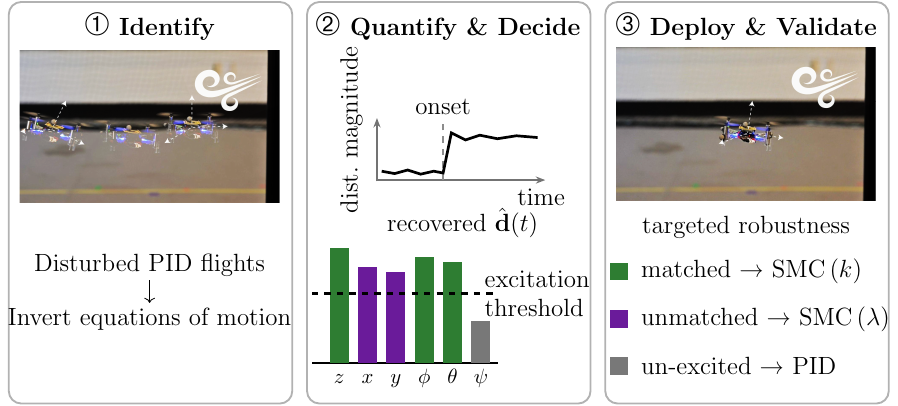}
    \caption{Selective SMC deployment workflow. PID hover flights under disturbance are inverted through the equations of motion~\ding{192}. The recovered per-axis disturbance $\hat{\mathbf d}(t)$ is scored by its excitation $\hat\delta$ against a threshold from nominal hovers to identify the excited axes~\ding{193}. Excited axes receive SMC, tuning the switching gain $k$ on matched axes and the surface slope $\lambda$ on unmatched axes, while the remaining axes retain PID~\ding{194}.
    \label{fig:cover_graphical}}
\vspace{-6mm}
\end{figure}

To overcome these linear limits, alternative robust control schemes exist, but each carries requirements on onboard resources, sensing fidelity, or modeling that can be difficult to meet on lightweight quadrotors. Estimate-and-cancel schemes, such as Incremental Nonlinear Dynamic Inversion (INDI)~\cite{SmeurINDIAttitude}, rely on high-rate angular acceleration feedback and precise control-effectiveness models, both of which are degraded onboard, where differentiated gyro signals are noisy and models only approximate. In disturbance and extended-state observers (DOBs/ESOs)~\cite{AhmedSMCDisturbanceObserver, WangExtendedStateObserver, CastilloDisturbanceObserver}, high observer gains amplify high-frequency IMU measurement noise and introduce phase lag through the low-pass filtering they typically require, while lower gains track fast transient disturbances more slowly. Adaptive schemes~\cite{MaFlatnessAdaptiveSMC, MeneboNeuroFuzzyASMC} require persistent excitation to guarantee parameter convergence; step disturbances at hover, such as an attached payload, provide little excitation, which can leave adaptation slow or prone to drift. Learning-based methods have shown promise in handling unmodeled dynamics~\cite{ShiNeuralLanderLearning, BansalQuadDynamicsLearning, CheeUncertainty}, but typically lack deterministic error bounds unless coupled with model predictive architectures~\cite{SinhaMatchedUnmatched, PanOnlineAdaptationWind}, whose real-time optimization remains considerably more expensive than a fixed control law.

Bound-based robust schemes are far less demanding to implement onboard: sliding mode control (SMC) requires only the tracking error, its derivative, and a disturbance bound, reducing online evaluation to a few algebraic operations per axis per cycle~\cite{SlotineNLControl}. Yet standard quadrotor SMC adds switching terms to every axis, including the lateral ones the vehicle cannot directly actuate, where a switching term does not improve rejection, adds parameters to tune, and can degrade tracking when switching terms in the position and attitude loops act on the same lateral force~\cite{YazdanshenasQuaternionSMC, GomieroSMCCascade, MeneboNeuroFuzzyASMC}. Moreover, such work is largely evaluated in continuous-time simulation under ideal state feedback~\cite{SalazarSimSMCQuad, GomieroSMCCascade}, leaving the discretization, chattering, and compute trade-offs of real hardware untested. Existing quadrotor SMC work does not use flight data to decide which axes need a switching term, which parameter to tune on each, or what it costs onboard.

Observer- and inversion-based methods estimate the disturbance online to cancel it on every axis. We instead measure the disturbance once, from PID hover flight logs, and our selective approach places a switching term only where a disturbance is both present and actuator-reachable (Fig.~\ref{fig:cover_graphical}). In a nutshell, the key intuition is that robustness is useful only when the controller has the authority to act.

\textbf{Contributions:} This work turns the matched and unmatched distinction into a practical procedure for deploying robust control. Specifically, the contributions are: 
\begin{enumerate}
    \item A per-axis deployment rule that identifies from PID hover flights where robust control belongs and which parameter to tune on each axis, built on actuation geometry and the gain response of boundary-layer SMC (Sec.~\ref{sec:selective_deployment}).
    \item Hardware validation under a slung payload and an oblique wind, showing that the switching gain reduces error only on matched axes while the surface slope governs unmatched ones, and that tracking improves on exactly the axes the rule selects (Secs.~\ref{subsec:ablation}--\ref{subsec:vicon_tracking}).
    \item An evaluation under fully onboard state estimation showing that selective SMC retains its advantage over INDI, at no added computational cost over the built-in PID (Secs.~\ref{subsec:optical_flow_exp}--\ref{subsec:compute}).
    \vspace{-1mm}
\end{enumerate}

\section{Preliminaries}
\subsection{Quadrotor Dynamic Model}
We use the standard rigid-body quadrotor model~\cite{MahonyMAVGeneral}, with world frame $\mathbf{W}$ ($z$ up), body frame $\mathbf{B}$, position $\mathbf{p} = [x, y, z]^\top$, Euler angles $\boldsymbol{\eta} = [\phi, \theta, \psi]^\top$, and body-to-world rotation
$\mathbf{R}(\boldsymbol{\eta})\in SO(3)$. Newton's and Euler's equations read
\begin{align}
    m\ddot{\mathbf{p}} &= m\mathbf{g} + \mathbf{R}(\boldsymbol{\eta})\mathbf{F} + \mathbf{d}_p,
    \label{eqn:trans-dyn}\\
    \mathbf{I}\dot{\boldsymbol{\omega}} &= \boldsymbol{\tau}
        - \boldsymbol{\omega}\times\mathbf{I}\boldsymbol{\omega} + \mathbf{d}_\omega,
    \label{eqn:rot-dyn}
\end{align}
with mass $m$, $\mathbf{g}=[0,0,-g]^\top$, body-frame thrust $\mathbf{F}=[0,0,F_z]^\top$, inertia $\mathbf{I}=\mathrm{diag}(I_{xx},I_{yy},I_{zz})$, body rates $\boldsymbol{\omega}$, and control torques $\boldsymbol{\tau}=[\tau_\phi,\tau_\theta,\tau_\psi]^\top$. Here $\mathbf{d}_p$ is a force disturbance and $\mathbf{d}_\omega$ a moment disturbance, both in $\mathbb{R}^3$, and we treat them separately, so each classification below is a split within $\mathbb{R}^3$ rather than of the full wrench. We assume that they are bounded, with $\|\mathbf{d}_p\|\le\delta_p$ and $\|\mathbf{d}_\omega\|\le\delta_\omega$, where $\delta_p$ and $\delta_\omega$ are positive constants. The bound values are not known \textit{a priori}, but are estimated from disturbed hovers through the excitation metric of Sec.~\ref{subsec:exc_thresh}. 

\subsection{Disturbance Classification}
\label{subsec:matched_unmatched}

Whether a controller can reject a disturbance or only attenuate it depends on how much of the disturbance lies in the directions its actuators reach. We define the span $\operatorname{range}(\mathbf{B}_p)\subseteq\mathbb{R}^3$ of the input map $\mathbf{B}_p(\boldsymbol\eta)$ from control inputs to translational force as the matched subspace, and its orthogonal complement as the unmatched subspace. With the orthogonal projector
\begin{equation}
    \mathbf{P}_m = \mathbf{B}_p(\mathbf{B}_p^\top\mathbf{B}_p)^{-1}\mathbf{B}_p^\top, 
    \label{eqn:pos_proj}
\end{equation}
any force disturbance decomposes into matched 
($\mathbf{d}_m=\mathbf{P}_m\mathbf{d}_p$) and unmatched 
($\mathbf{d}_u=(\mathbf{I}-\mathbf{P}_m)\mathbf{d}_p$) components. Actuators reject $\mathbf{d}_m$ directly, whereas $\mathbf{d}_u$ can only be attenuated through changes in attitude. Fully actuated vehicles ($\operatorname{range}(\mathbf{B}_p)=\mathbb{R}^3$) lack an unmatched subspace. In contrast, underactuated platforms like quadrotors, whose thrust spans only the body $z$-axis (Sec.~\ref{subsec:spec_quads}), inherently retain an unmatched component. 

We use matched in its classical sense, with respect to the physical inputs $F_z$ and $\boldsymbol\tau$. In a cascade, the outer position loop does not command these inputs directly. It commands a tilt, which the attitude loop then realizes. With respect to that commanded tilt, a lateral force could be considered matched. The distinction still matters because the tilt is realized only through the attitude loop, with finite bandwidth and a tilt limit. This realization, not the outer loop's gain, bounds how well a lateral disturbance can be opposed (Sec.~\ref{subsec:scalar_to_cascade}).

\subsection{Specialization to Quadrotor}
\label{subsec:spec_quads}
A quadrotor produces translational force only along its thrust axis, so $\mathbf{B}_p = \mathbf{R}\mathbf{e}_3$ is a single unit vector, and the projector~\eqref{eqn:pos_proj} reduces to a component split: the part of $\mathbf{d}_p$ along the thrust axis is matched, and the part in the perpendicular plane is unmatched. In the body frame, where thrust is always $\mathbf{e}_3$,
\begin{equation}
    d_m = \big|d_z^{b}\big|, \qquad d_u = \sqrt{(d_x^{b})^2 + (d_y^{b})^2},
\label{eqn:split_quad}
\end{equation}
with $\mathbf{d}^{b} = \mathbf{R}^\top \mathbf{d}_p = [d_x^{b}, d_y^{b}, d_z^{b}]^\top$ the force disturbance of~\eqref{eqn:trans-dyn} expressed in the body frame. As the body frame rotates with the vehicle, this holds at any attitude and needs no small-angle assumption. At hover, a vertical disturbance such as an attached payload is therefore matched and can be rejected, while a lateral one such as oblique wind is unmatched, since the quadrotor lacks lateral actuation and can only attenuate it by tilting to redirect thrust.

We compute both this split and the per-axis excitation metric (Sec.~\ref{subsec:exc_thresh}) in the body frame, in which the accelerometer and thrust are natively expressed. This avoids a rotation by the estimated attitude, through which any attitude error would project gravity into the lateral components. Near hover, with small tilt and yaw regulated to its setpoint, each body axis remains closely aligned with the corresponding controlled axis.
 
Control torques, by contrast, act about all three body axes independently ($\mathbf{B}_\omega=\mathbf{I}$), so every disturbance moment $\mathbf{d}_\omega$ is matched by construction (Table~\ref{tab:disturbance_classification}).

\begin{table}
\centering
\caption{\small{Per-axis actuation geometry}}
\setlength{\tabcolsep}{4pt}
\begin{tabular}{lcc}
\hline
\textbf{Axis} & \textbf{Primary control input} & \textbf{Channel structure} \\
\hline
$z$          & Thrust ($F_z$)              & Matched \\
$x, y$       & Attitude (indirect)         & Unmatched \\
$\phi,\theta,\psi$& Torque ($\tau_\phi,\tau_\theta,\tau_\psi$) & Matched \\
\hline
\end{tabular}
\label{tab:disturbance_classification}
\vspace{-5mm}
\end{table}

\section{Sliding Mode Control and its Guarantees}
\label{sec:control}
\subsection{Boundary-layer Sliding Mode Control}
\label{subsec:smc_framework}
Each axis is regulated independently, so we state the law for the scalar subsystem
\begin{equation}
    \ddot{\xi} = f(\xi,\dot\xi) + b(\xi,\dot\xi)\,u + d,
    \label{eqn:ctrl_affine_sys}
\end{equation}
where $\xi$ is the controlled variable, $f$ the drift dynamics, $u$ the control input, and $d$ a bounded disturbance. The control effectiveness $b(\xi,\dot\xi)>0$ is the gain from input to acceleration. Because the SMC issues setpoints to the built-in inner loop rather than forces or torques, we use a nominal value $b_{nom}$, the effective gain of the inner loop it commands (velocity on the position axes, body rate on the attitude axes), identified from its step response. For tracking error $e=\xi-\xi_{des}$ and surface slope $\lambda>0$, we define the sliding surface and control law
\begin{equation}
s = \dot e + \lambda e, \qquad u = u_{eq} + u_{sw},
\label{eqn:sliding_surf}
\end{equation}
so that on the manifold $\mathcal{S}=\{s=0\}$ the error decays exponentially at rate $\lambda$. The equivalent control is implemented as a PD law, $u_{eq} = -K_{p}^{eq}e - K_{d}^{eq}\dot e$, tuned for nominal hover, and $\lambda$ therefore enters only through the switching term. Setting $k=0$ leaves exactly this PD law, which we use as the reference for isolating the switching term's contribution. This reference is distinct from the Crazyflie's built-in controller, a cascaded PID with its own gains and integral action, which we report as a separate baseline. The switching term is smoothed over a boundary layer of thickness $\varphi>0$ to suppress chattering~\cite{SlotineNLControl},
\begin{equation}
    u_{sw} =
    \begin{cases}
        -k\,\mathrm{sgn}(s) & |s|>\varphi,\\[2pt]
        -k\,s/\varphi & |s|\le\varphi,
    \end{cases}
    \label{eqn:u_sw}
\end{equation}
with switching gain $k>0$. On the attitude axes, the error derivatives are taken directly from the gyroscope, since near hover the Euler-rate and body-rate maps agree to second order~\cite{MahonyMAVGeneral}, giving $\dot e_\phi\approx\omega_x$ and $\dot e_\theta\approx-\omega_y$, the sign reflecting the Crazyflie's inverted pitch convention.

\begin{figure*}[bth]
    \centering
    \begin{subfigure}[t]{0.32\textwidth}
        \centering
        \includegraphics[width=\linewidth]{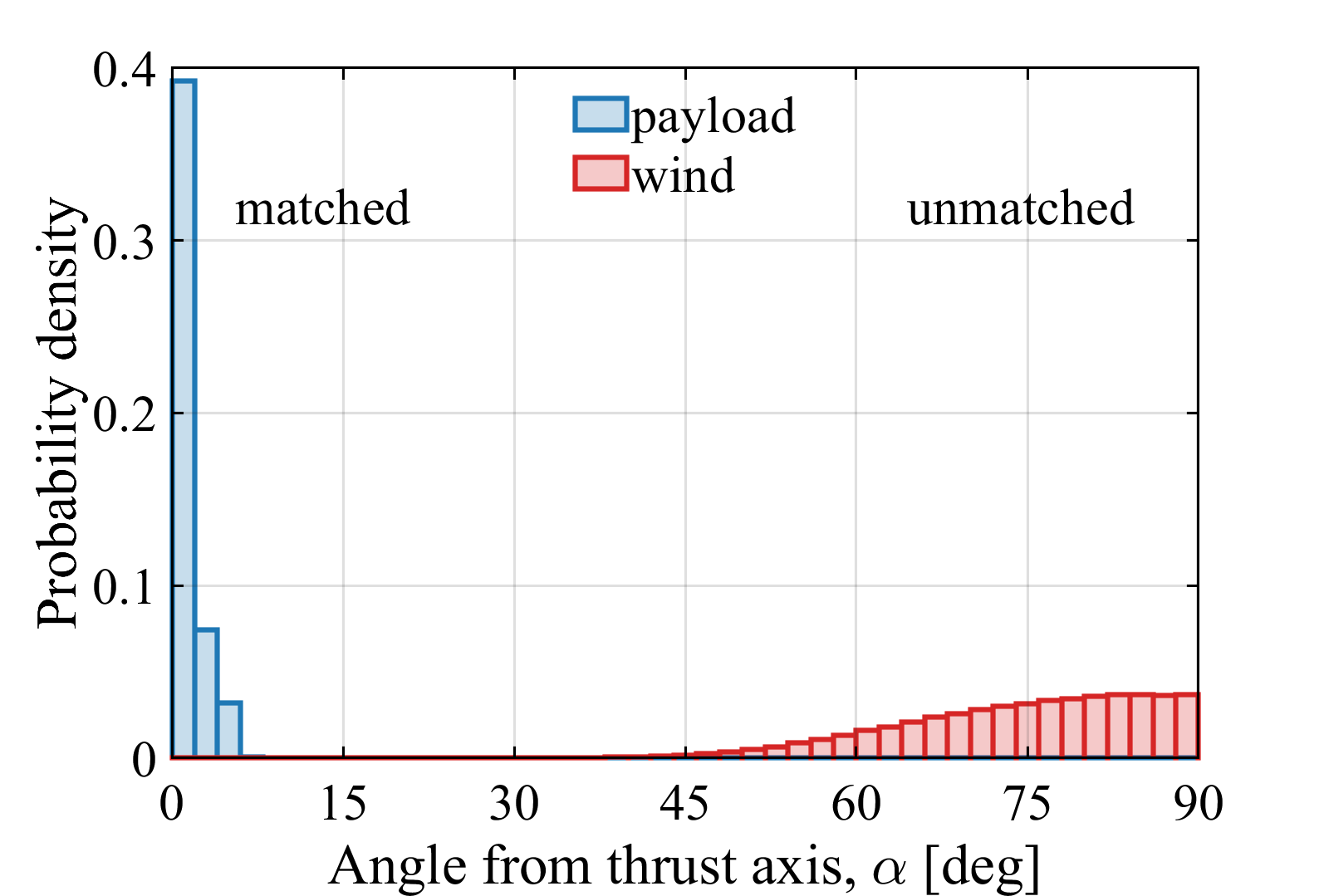}
        \caption{Measured direction from thrust axis}
        \label{fig:matched_unmatched_angle}
    \end{subfigure}
    \hfill
    \begin{subfigure}[t]{0.32\textwidth}
        \centering
        \includegraphics[width=\linewidth]{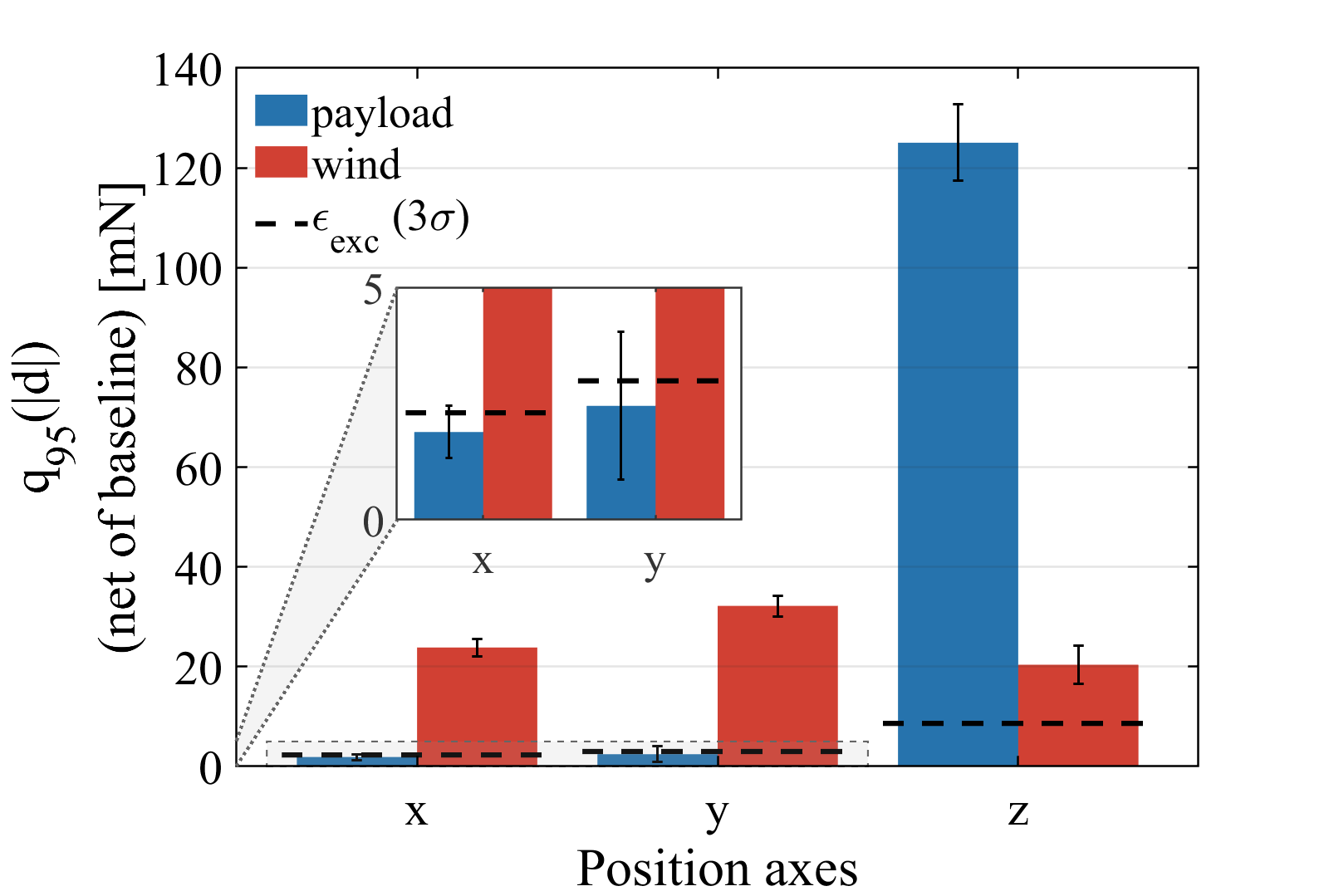}
        \caption{Translational excitation}
        \label{fig:position_bars}
    \end{subfigure}
    \hfill
    \begin{subfigure}[t]{0.32\textwidth}
        \centering
        \includegraphics[width=\linewidth]{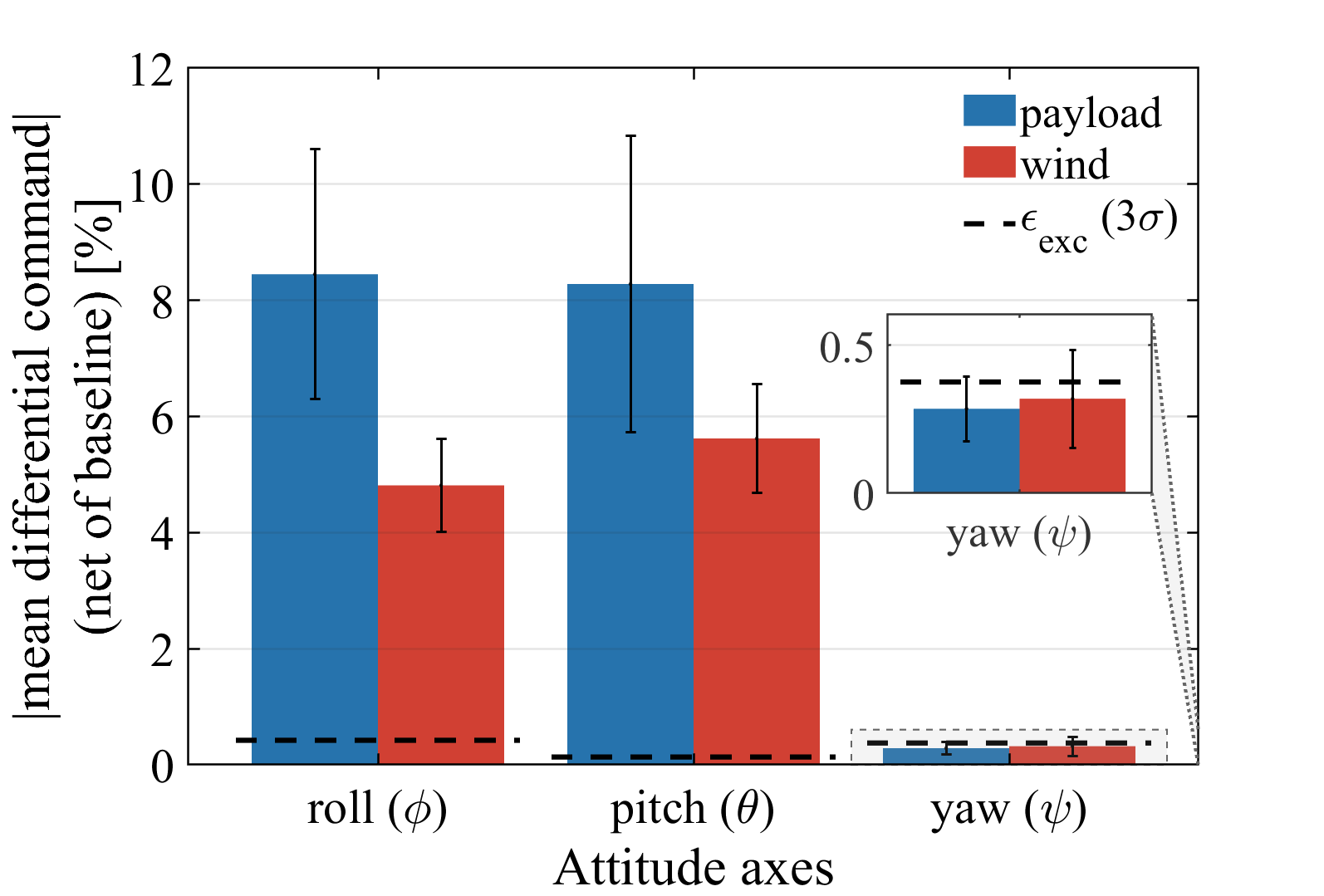}
        \caption{Attitude excitation}
        \label{fig:attitude_bars}
    \end{subfigure}
    \caption{\footnotesize Disturbance classification from flight data. (a) Angle of the recovered disturbance relative to the thrust axis, clustering near $0^\circ$ for matched disturbances (payload) and near $90^\circ$ for unmatched disturbances (wind). (b, c) Per-axis excitation $\hat\delta_i$~\eqref{eqn:delta_hat}, as force (mN) for position and as a percentage of the summed motor commands for attitude, compared against the excitation threshold $\varepsilon_{\mathrm{exc}}$ (dashed), with bars indicating mean $\pm$ std. Insets highlight less-affected axes ($x, y$ for position, yaw for attitude) below the threshold. Axes remaining below threshold retain PID control.}
    \label{fig:disturbance_classification}
    \vspace{-2mm}
\end{figure*}

\subsection{Guarantees provided by switching term}
\label{subsec:stability}
The analysis carries a single uncertainty, an additive modeling residual bounded by $\rho\ge|\Delta f|+|\Delta b|\,|u_{eq}|$, where $\Delta f$ and $\Delta b$ are the drift and effectiveness errors, respectively. We measure $\rho$ directly as the baseline residual floor from flight data (Sec.~\ref{sec:selective_deployment}), so effectiveness uncertainty is absorbed into a quantity we observe. 

\begin{theorem}[Bounded disturbance rejection~\cite{SlotineNLControl}]
\label{thm:bounded_dist}
For~\eqref{eqn:ctrl_affine_sys} with disturbance bound $|d|\le\delta$, the boundary-layer SMC~\eqref{eqn:u_sw} with $k>(\rho+\delta)/b_{nom}$ achieves a steady-state tracking error
\begin{equation}
|e|_{ss} \le \frac{\varphi(\rho+\delta)}{\lambda\,k\,b_{nom}}.
\label{eqn:steady_state_bound}
\end{equation}
\end{theorem}
The bound follows from the standard boundary-layer argument~\cite{SlotineNLControl}, in which a switching gain large enough to dominate the disturbance drives the sliding variable into the layer in finite time, and the steady-state error inside the layer scales inversely with $k$.

\subsection{From single axis to the cascade}
\label{subsec:scalar_to_cascade}
The results above treat one axis in isolation, whereas the deployed controller acts through the cascade, issuing velocity and attitude-rate setpoints rather than forces and torques. Two properties of the cascade let the single-axis results carry over. First, the inner attitude loop is much faster than the outer position loop, so the attitude has effectively settled on the timescale over which position evolves, and each outer axis behaves as the scalar system~\eqref{eqn:ctrl_affine_sys}. Second, position is controlled only by commanding a tilt, so the coupling runs one way. The outer loop sets the inner loop's reference, but the inner loop's error dynamics do not depend on position. Each loop is therefore ultimately bounded by Theorem~\ref{thm:bounded_dist}. The attitude error enters the translational dynamics as an additional bounded disturbance, so Theorem~\ref{thm:bounded_dist} also applies to the outer loop with that disturbance added to its bound. Because~\eqref{eqn:steady_state_bound} is linear in the disturbance bound, the outer-loop error is its own per-axis bound plus a residual from the attitude error.
The outer loop thus pays a small penalty for imperfect attitude tracking, and it shrinks as the attitude loop tracks more tightly.

\begin{proposition}[Disturbance-dependent performance]
\label{prop:matched_unmatched}
Under SMC~\eqref{eqn:u_sw}, a matched disturbance is driven to $|e|_{ss}\to0$ as $k\to\infty$, whereas an unmatched disturbance leaves a floor independent of $k$, reducible only through the surface slope $\lambda$.
\end{proposition}
The matched case follows from~\eqref{eqn:steady_state_bound}. For an unmatched disturbance, the correction must pass through the rate- and tilt-limited attitude path, described above, so added gain has no effect, while $\lambda$, which shapes the correction, still does. 

\section{Selective Deployment}
\label{sec:selective_deployment}
In this section, we state our main contribution, the per-axis deployment rule. Geometry fixes which axes a disturbance can be opposed on, and Proposition~\ref{prop:matched_unmatched} fixes when opposing it reduces the error and which parameter to tune. Neither can say which axes a particular disturbance excites, since both are fixed before the vehicle flies. We recover that from flight data, and combine all three into the rule (Algorithm~\ref{alg:deploy}).

\subsection{Recovering the disturbance}
\label{subsec:recov_dist}
The onboard accelerometer measures the specific force $\mathbf{a}_s=\mathbf{R}^\top(\ddot{\mathbf p}-\mathbf g)$ in the body frame, so inverting the translational dynamics~\eqref{eqn:trans-dyn} gives the lumped disturbance directly in the body frame,
\begin{equation}
    \hat{\mathbf{d}}^{b} = m\,\mathbf{a}_s - \mathbf{F},
    \label{eqn:lumped_dist_pos}
\end{equation}
where $\mathbf{F}$ is reconstructed from the logged motor commands through a per-motor thrust map identified from hover flights with known added masses. Every term on the right is therefore measured or known. Because the applied thrust is taken from the log rather than modeled, the controller's own response appears in $\mathbf{F}$, and $\hat{\mathbf{d}}^{b}$ is the residual between the measured and the nominal-model acceleration. Thrust-map error and sensor noise leave a nonzero residual even without a disturbance, which is the modeling residual $\rho$ of Sec.~\ref{subsec:stability}; we measure it from nominal hovers in Sec.~\ref{subsec:exc_thresh}. On the attitude axes, the steady control effort holding the vehicle level absorbs a constant disturbance moment, which we read from the differential thrust the mixer sustains to oppose it. Roll and pitch come from opposing-arm command differences and yaw from the reaction-torque difference between clockwise and counter-clockwise rotor pairs,
\begin{equation}
\begin{aligned}
    \hat{d}_\phi = \tfrac{(f_1+f_4)-(f_2+f_3)}{F_\Sigma} - \bar d_\phi,\\[3pt]
    \hat{d}_\theta = \tfrac{(f_1+f_2)-(f_3+f_4)}{F_\Sigma} - \bar d_\theta,\\[3pt]
    \hat{d}_\psi = \tfrac{(f_1+f_3)-(f_2+f_4)}{F_\Sigma} - \bar d_\psi,
\end{aligned}
\label{eqn:diff_thrust}
\end{equation}
with $f_{i}$ the logged motor commands, $F_\Sigma = \sum_i f_{i}$ their sum, and $\bar d_{(\cdot)}$ the mean trim over the nominal hovers. The yaw expression assumes rotor reaction torque proportional to thrust. The resulting sustained command differential reflects the disturbance moment that the controller rejects

Classification uses flights with the disturbance present from takeoff, so the steady window is uniformly disturbed. Flights with the payload attached during hover show the same picture over time. The payload steps onto the thrust axis and stays there, while its lateral component appears only briefly at attachment and then decays, so the lateral axes stay below the excitation threshold.

Applying~\eqref{eqn:split_quad} to the recovered disturbance shows that it separates cleanly by direction, with the payload concentrating near the thrust axis (matched) and the wind in the perpendicular plane (unmatched), confirming the geometric classification from measurement (Fig.~\ref{fig:matched_unmatched_angle}).

\subsection{Excitation metric and threshold}
\label{subsec:exc_thresh}
For any flight, we summarize the recovered disturbance on each axis over the steady hover window, after removing the mean offset measured on nominal hovers. On the position axes, we use the 95th percentile of the force magnitude, so that isolated transients do not inflate the estimate. On the attitude axes, a constant moment appears as a sustained mixer differential; we use the magnitude of its mean as a percentage of the summed commands,
\begin{equation}
        \hat\delta_i =
    \begin{cases}
        \mathrm{q}_{95}\!\big(|\hat d^{b}_{i}-\bar d^{b}_{i}|\big),
            & i \in \{x,y,z\}\ \text{from~\eqref{eqn:lumped_dist_pos}},\\[4pt]
        \big|\mathrm{mean}(\hat d_{i})\big|,
            & i \in \{\phi,\theta,\psi\}\ \text{from~\eqref{eqn:diff_thrust}},
    \end{cases}
    \label{eqn:delta_hat}
\end{equation}
with $\bar d^{b}_{i}$ the mean nominal-hover residual. For the gain rule, we convert these to accelerations, dividing the position entries by $m$ and converting the attitude entries to angular acceleration, $\hat\delta_i\, T\, \ell / I_{ii}$, with $T$ the collective thrust from the thrust map and $\ell$ the effective moment arm.

Inverting the dynamics returns a nonzero residual even without a disturbance because of model mismatch and sensor noise. We therefore evaluate the metric on two sets of flights. Over $N_{1}=15$ nominal hovers, its mean gives the floor $\rho_{i}$, and its standard deviation gives $\sigma_{\rho,i}$. Over $N_{2}=15$ hovers under the applied disturbance, it gives $\hat\delta_{i}$.

An axis is classified as excited when $\hat\delta_{i}$ exceeds the floor by more than $n$ standard deviations,
\begin{equation}
\hat\delta_i > \varepsilon_{\mathrm{exc},i}, \qquad
\varepsilon_{\mathrm{exc},i} = \rho_i + n\,\sigma_{\rho,i}.
\label{eqn:exc_threshold}
\end{equation}
with $n = 3$, the standard three-sigma choice that baseline variation alone rarely exceeds. For the two disturbances tested here, excited residuals clear the threshold by a wide margin in practice, so the classification is invariant across any choice of $n\in[1,5]$. The threshold matters most for weaker disturbances whose residual approaches the noise floor, where it prevents deploying switching terms on axes whose apparent excitation is model mismatch. 

\begin{figure*}[tbh]
\centerline{\includegraphics[scale=0.7]{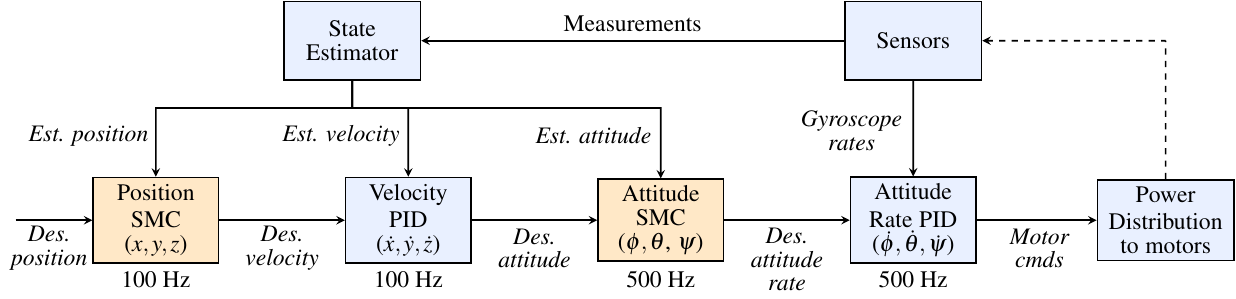}}
\caption{Selective SMC deployed onto the Crazyflie's cascaded control architecture. Yellow marks loops that can be replaced by SMC, and blue marks loops that always retain PID. Position-level SMC issues velocity setpoints at 100\,\si{Hz}, and attitude-level SMC issues body-rate setpoints at 500\,\si{Hz}, both to the built-in inner loops. Algorithm~\ref{alg:deploy} selects which yellow loops are switched to SMC for a given disturbance.}
\label{fig:cascaded_controller_flowchart}
\vspace{-2mm}
\end{figure*}

Applying the rule to the two test disturbances, each characterized over the $N_{2}$ disturbed hovers, shows how sharply it discriminates. A slung payload exceeds the threshold in altitude, roll, and pitch but leaves the lateral axes unexcited (Figs.~\ref{fig:position_bars} and~\ref{fig:attitude_bars}). Oblique wind exceeds the threshold on the lateral axes, roll, pitch, and altitude, because holding position against the lateral load forces the vehicle to tilt, projecting part of the disturbance onto the thrust axis. Sec.~\ref{subsec:ablation} tests the classification through gain sweeps on the selected and excluded axes.

\subsection{Deployment and tuning}
\label{subsec:deploy_and_tune}
The rule combines the two tests. An axis receives a switching term only if it is excited; its actuation class (Table~\ref{tab:disturbance_classification}) then sets whether the switching gain $k$ (matched) or the surface slope $\lambda$ (unmatched) is tuned, per Proposition~\ref{prop:matched_unmatched}. Every axis below threshold retains PID. Algorithm~\ref{alg:deploy} gives the full procedure.

One constraint follows from the cascade. Since position is controlled only through attitude, a lateral axis and the attitude axes that tilt the vehicle along it form a single path. When a lateral axis is excited, the switching term is placed at the position level, and the attitude axes on that path retain PID, since switching terms at both levels interact through the tilt coupling and degrade tracking.

Our controller augments the Crazyflie's cascaded controller architecture~\cite{bitcrazeControllersCrazyflie} (Fig.~\ref{fig:cascaded_controller_flowchart}) with the selected switching terms, replacing the built-in PID on the deployed loops while retaining it elsewhere. 

\begin{algorithm}[tb]
\caption{Selective SMC deployment and per-axis tuning}
\begin{algorithmic}[1]
\STATE Fly $N_{1}$ nominal hovers (PID); evaluate the metric~\eqref{eqn:delta_hat} per axis
\STATE $\rho_i \gets \mathrm{mean}_{N_1}$, \; $\sigma_{\rho,i}\gets\mathrm{std}_{N_1}$; \; set $\varepsilon_{\mathrm{exc},i}=\rho_i+3\sigma_{\rho,i}$
\STATE Fly $N_{2}$ disturbed hovers (PID); compute $\hat\delta_i$ per~\eqref{eqn:delta_hat}
\FOR{each axis $i$}
    \IF{$\hat\delta_i \le \varepsilon_{\mathrm{exc},i}$}
        \STATE retain PID \COMMENT{below threshold}
    \ELSIF{axis $i$ matched (Table~\ref{tab:disturbance_classification})}
        \STATE deploy SMC, set $k>(\rho_i+\hat\delta_i)/b_{nom}$ \COMMENT{matched, Prop.~\ref{prop:matched_unmatched}}
    \ELSE
        \STATE deploy SMC, set $\lambda$ from convergence rate \COMMENT{unmatched, Prop.~\ref{prop:matched_unmatched}}
    \ENDIF
\ENDFOR
\STATE \textbf{if} a lateral axis is deployed, retain PID on its attitude path.
\end{algorithmic}
\label{alg:deploy}
\end{algorithm}

\section{Physical Experimental Results}
\label{sec:phys_exps}

Our experiments evaluate the selective deployment rule on hardware. Gain sweeps confirm the property it rests on: that the switching gain reduces error only on matched axes, while the surface slope governs the unmatched ones. Subsequent disturbance-rejection flight experiments under purely matched and mixed disturbances show that tracking improves on the axes the rule selects. Finally, a fully onboard comparison demonstrates that this benefit holds under imperfect state estimation. Tracking statistics use ten trials per controller configuration and three per gain value in the sweeps. 

For experiments, we use a Crazyflie 2.1 quadrotor (38.8\,\si{g} as flown, including markers) in an indoor motion capture arena. Twenty-four Vicon Valkyrie cameras provide 6-DOF pose estimation at 120\,\si{Hz} with $\sim$0.5\,\si{mm} position and $\leq$0.1$^\circ$ attitude accuracy. State feedback is relayed via Crazyradio~\cite{HonigFlightROS} at 100\,\si{Hz} to the onboard STM32F405 microcontroller (168\,\si{MHz}), which executes the cascaded position and attitude control of Sec.~\ref{subsec:deploy_and_tune}. The onboard extended Kalman filter fuses the motion-capture pose with IMU measurements and provides the velocity used by the position loop. The PID controller employs Crazyflie's default gains, and SMC deployment follows Algorithm~\ref{alg:deploy}. We exclude the initial takeoff phase from the performance evaluation.

\begin{figure}[tb]
    \centering
    \begin{subfigure}[t]{0.21\textwidth}
        \centering
        \includegraphics[width=\linewidth]{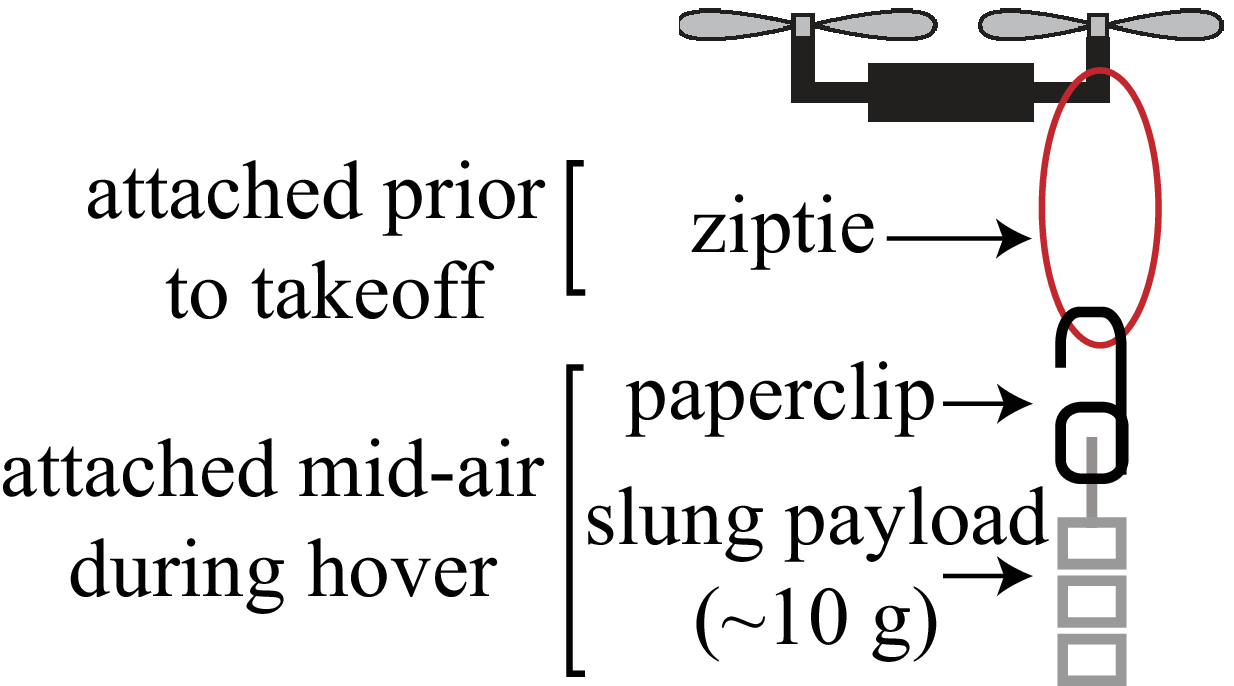}
        \caption{Slung payload (side-view)}
        \label{fig:payload_exp}
    \end{subfigure}
    \hspace{0.005\textwidth}
    \begin{subfigure}[t]{0.20\textwidth}
        \centering
        \includegraphics[width=\linewidth]{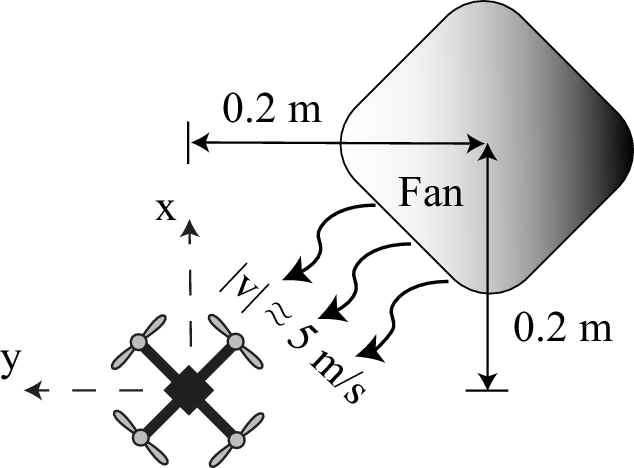}
        \caption{Oblique wind (top-view)}
        \label{fig:wind_exp}
    \end{subfigure}
    \caption{\footnotesize Setup for the physical experiments in Sec.~\ref{sec:phys_exps}.}
    \label{fig:phys_exps}
    \vspace{-4mm}
\end{figure}

\subsection{Gain sweeps}
\label{subsec:ablation}
This section tests the deployment rule's two premises: i) that the matched/unmatched classification predicts how each axis's error responds to the switching gain; and ii) that the switching term improves tracking only on axes that are both excited and matched. Both are tested by sweeping $k$ on one axis with all others fixed. Because $k=0$ leaves the equivalent PD alone, each sweep isolates the switching term's contribution.

\begin{figure*}[tbh]
  \centering
  \begin{subfigure}[t]{0.325\textwidth}
    \centering
    \includegraphics[width=\linewidth]{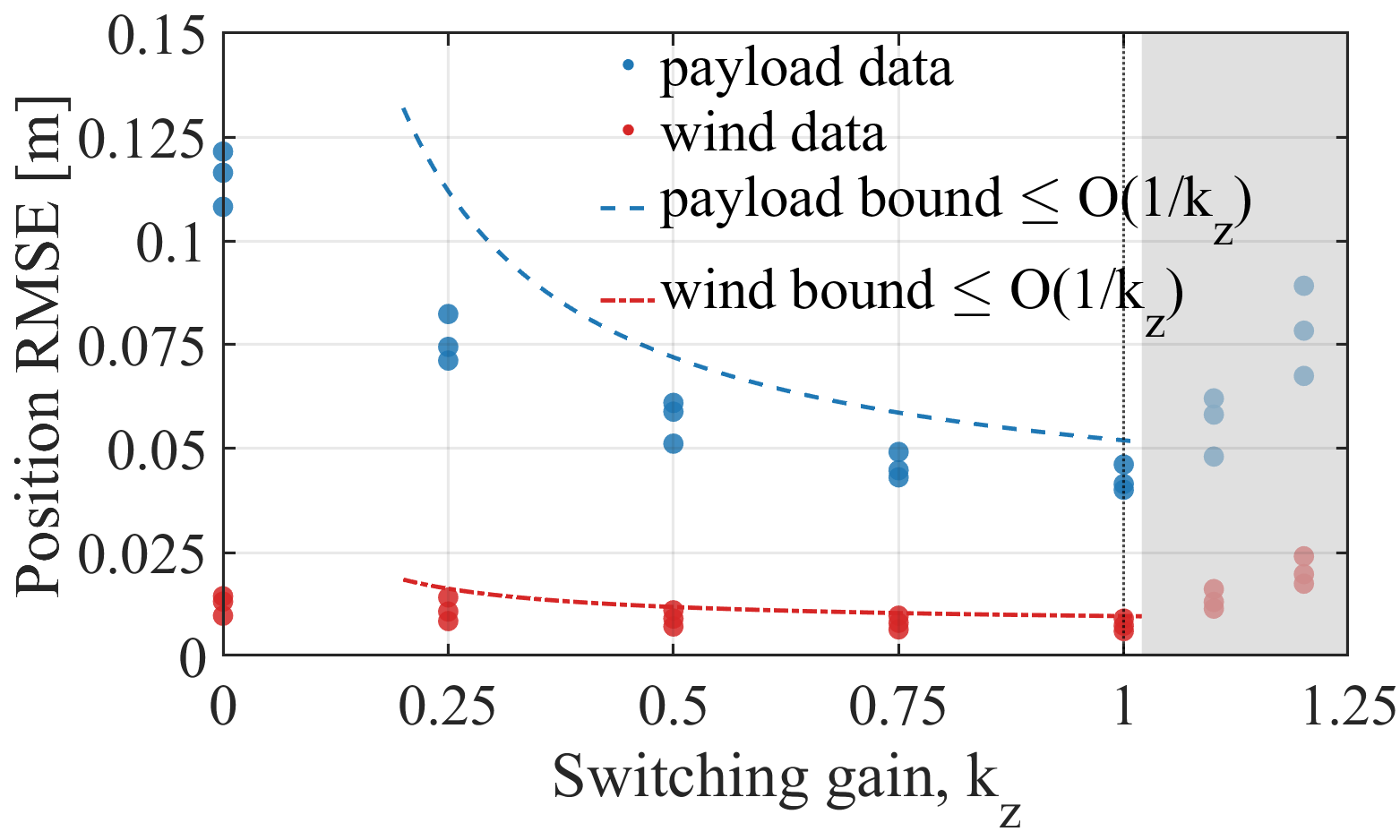}
    \caption{Matched ($z$) gain sweep}
    \label{fig:ksweep_pos_z}
  \end{subfigure}
  \hfill
  \begin{subfigure}[t]{0.325\textwidth}
    \centering
    \includegraphics[width=\linewidth]{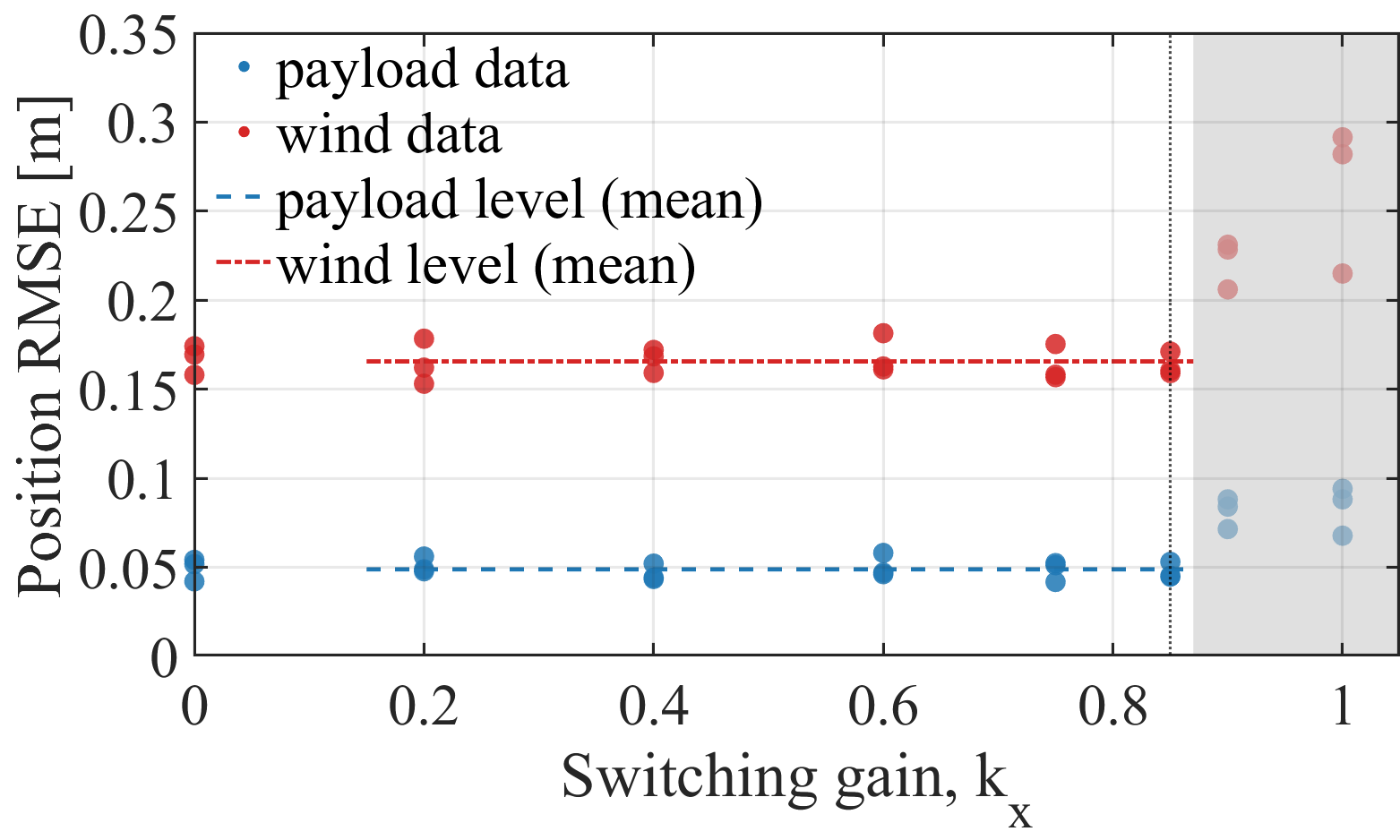}
    \caption{Unmatched ($x$) gain sweep}
    \label{fig:ksweep_pos_x}
  \end{subfigure}
  \hfill
  \begin{subfigure}[t]{0.325\textwidth}
    \centering
    \includegraphics[width=\linewidth]{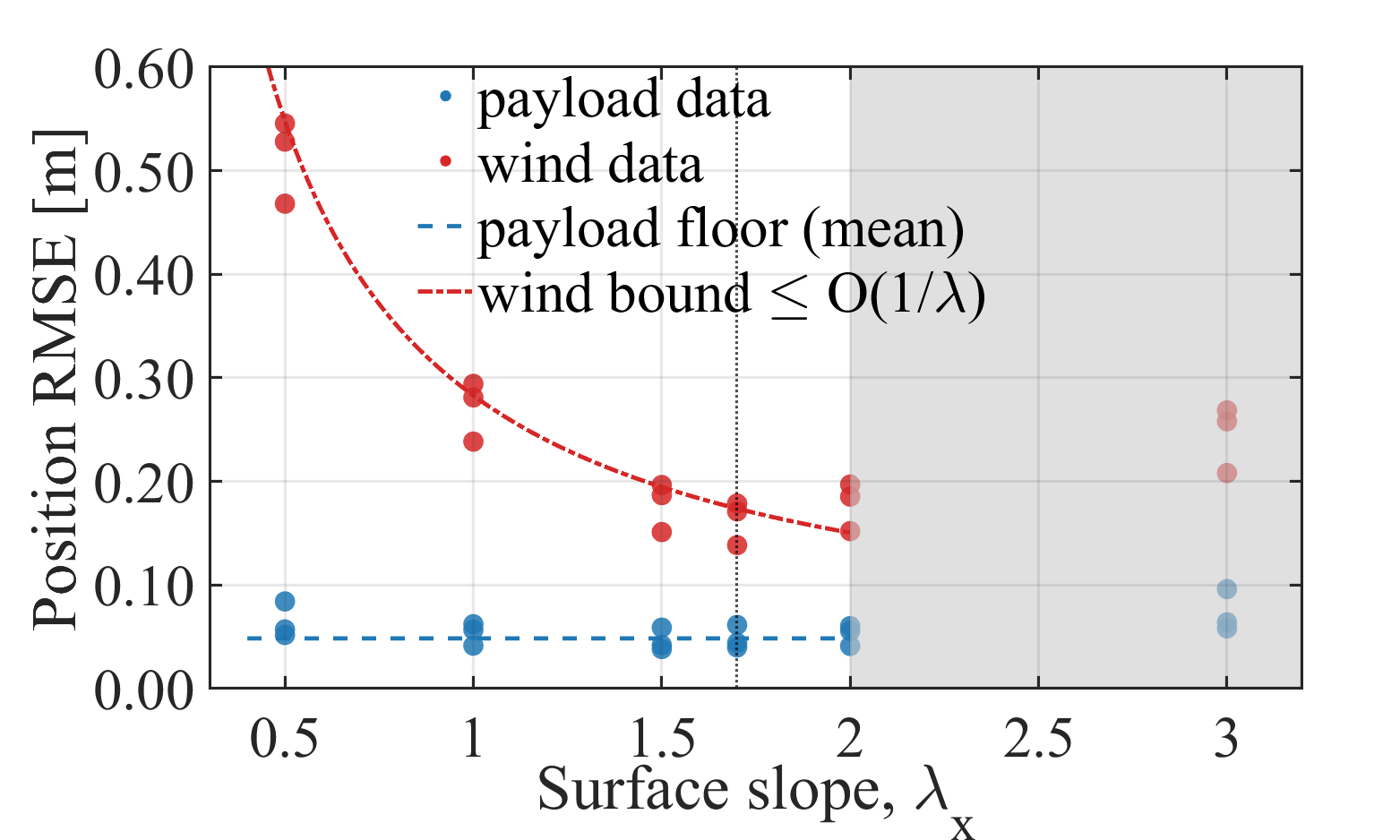}
    \caption{Unmatched ($x$) slope sweep}
    \label{fig:lambda_sweep_pos_x}
  \end{subfigure}
  \caption{\footnotesize Position-level results. (a, b) RMSE vs.\ switching gain $k$ for the payload (blue) and wind (red) disturbances, three trials each. Dashed lines mark the $\mathcal{O}(1/k)$ scaling of~\eqref{eqn:steady_state_bound} in (a) and the $k$-independent level in (b), drawn at the mean RMSE across the sweep, and the shaded band in (a, b) marks the chattering regime at high $k$. (c) Lateral RMSE vs.\ surface slope $\lambda$ under payload and wind at fixed $k$, three trials each. Dotted vertical lines mark the deployed $k$ in (a, b) and the deployed $\lambda$ in (c).}
  \label{fig:k_sweep_and_lambda_sweep}
  \vspace{-4mm}
\end{figure*}

The first premise is that the matched and unmatched classification predicts how error responds to gain. On altitude, a matched axis, the error falls as $\mathcal{O}(1/k)$ from the equivalent PD alone at $k=0$ and tracks the predicted bound~\eqref{eqn:steady_state_bound} until chattering takes over at high $k$ (Fig.~\ref{fig:ksweep_pos_z}), isolating the reduction to the switching term. Our methodology anticipates this before any tuning, since the classification marks altitude as matched. The payload disturbance, which sits almost entirely on the thrust axis, drives this reduction harder than the wind, which reaches that axis only through the vehicle's tilt, leaving less matched error for the switching term to compensate (cf. Figs.~\ref{fig:matched_unmatched_angle} and~\ref{fig:position_bars}).

The second premise is that the switching term helps only where the disturbance is both present and matched. The lateral axes under payload never cross the excitation threshold (Fig.~\ref{fig:position_bars}), and, as expected, raising $k$ leaves their error unchanged. Under wind, the lateral axes are excited, but the disturbance is unmatched, and $k$ again has no effect (Fig.~\ref{fig:ksweep_pos_x}). Raising $\lambda$ at fixed $k$ instead reduces lateral RMSE under wind from 0.51\,\si{m} at $\lambda=0.5$ to 0.16\,\si{m} at $\lambda=1.7$, while under payload, where the lateral axes are not excited, it has little effect (Fig.~\ref{fig:lambda_sweep_pos_x}), as Proposition~\ref{prop:matched_unmatched} predicts. Our approach thus identifies in advance the parameter worth tuning on each axis, a choice that uniform deployment leaves to trial and error. 

\subsection{Motion Capture Tracking Performance}
\label{subsec:vicon_tracking}
The payload excites only matched axes, while oblique wind excites both matched and unmatched axes, so the two experiments test whether the rule deploys robustness differently in each case. 

\subsubsection{Slung Payload Experiments}
\label{subsec:slung_payload_exp}
A 11.5\,\si{g} slung payload, approximately~\emph{one-third of the vehicle's mass}, is attached to the Crazyflie during hover, ${\sim}20$\,\si{s} after it reaches the commanded height. The payload is suspended by a paperclip and ziptie (Fig.~\ref{fig:payload_exp}), free to swing in $x$ and $y$ directions, loading the vehicle along thrust (added mass) and in roll and pitch (pendulum-induced moments). We evaluated the controller configurations in Table~\ref{tab:payload_controllers} across 10 trials each.

\begin{table}[tbh]
\centering
\caption{\small Controller assignments for payload experiments}
\setlength{\tabcolsep}{3pt}
\begin{tabular}{lccc}
\hline
\textbf{Controller config.} & \textbf{$x$-$y$ Pos.} & \textbf{$z$ Pos.} & \textbf{Attitude ($\phi,\theta,\psi$)} \\
\hline
PID-z/PID-att   & PID & PID & PID \\
SMC-z/PID-att   & PID & SMC & PID \\
SMC-z/SMC-att   & PID & SMC & SMC ($\phi,\theta$), PID ($\psi$) \\  
\hline
\end{tabular}
\label{tab:payload_controllers}
\end{table}

\begin{figure}[tb]
    \centering
    \includegraphics[width=1\linewidth]{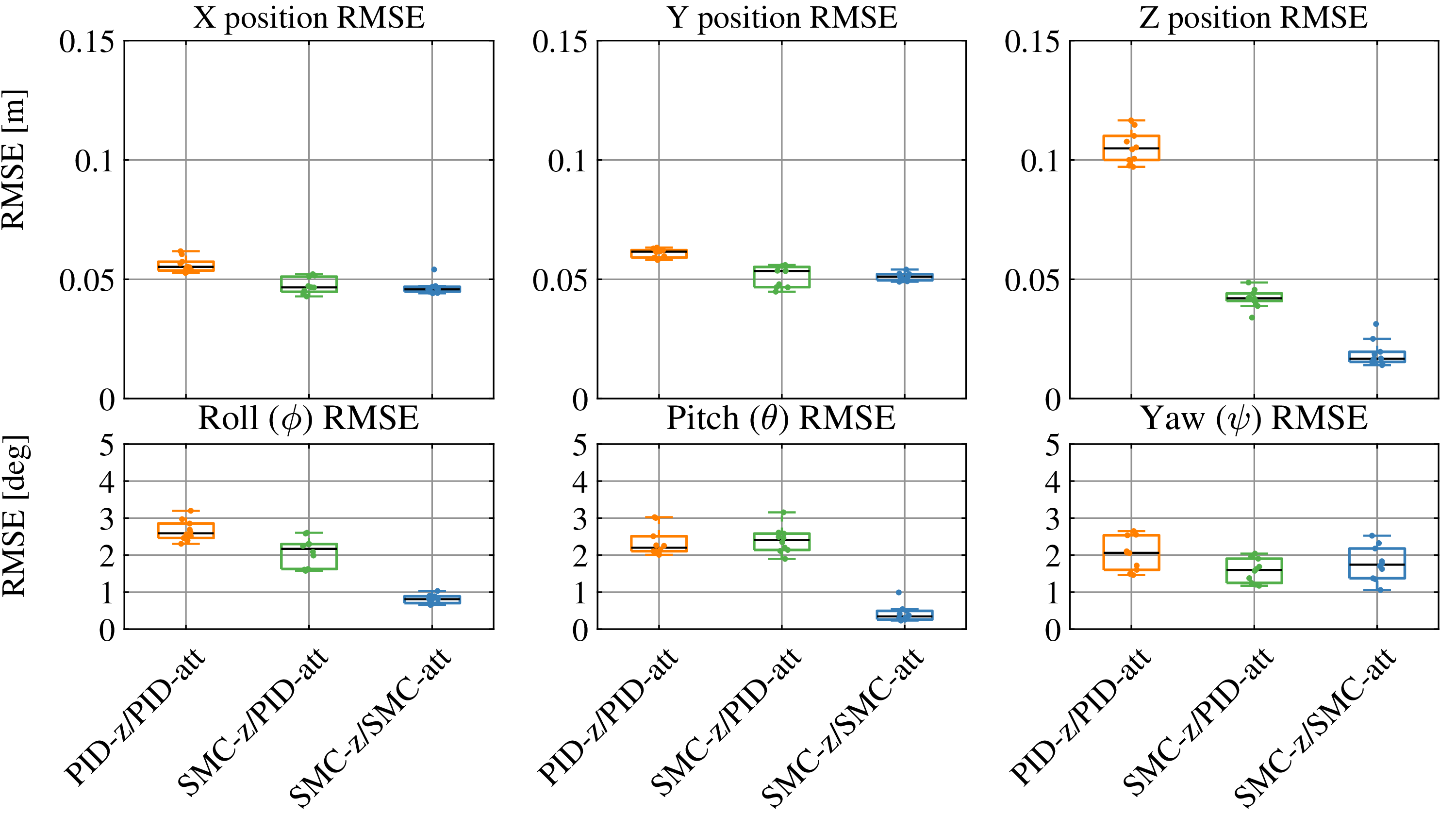}
    \caption{\footnotesize
    Per-axis RMSE for slung payload experiments (Sec.~\ref{subsec:slung_payload_exp}), ten trials per controller configuration. Dots show individual trials, black lines medians, and bars the 25th--75th percentiles.
    \label{fig:payload_data_vicon_rmse}}
    \vspace{-5mm}
\end{figure}

SMC on the matched axes gives the lowest altitude error (Fig.~\ref{fig:payload_data_vicon_rmse}). Against the built-in PID, altitude RMSE drops 82\% ($0.105\to0.019$\,\si{m}), roll 70\% ($2.66^\circ\!\to\!0.81^\circ$), and pitch 82\% ($2.35^\circ\!\to\!0.42^\circ$). The altitude sweep (Sec.~\ref{subsec:ablation}) shows that the switching term accounts for the entire altitude reduction, consistent with the $\mathcal{O}(1/k)$ matched rejection of~\eqref{eqn:steady_state_bound}. Our rule left the lateral axes and yaw on PID, and the gains appear exactly on the axes it selected.

\subsubsection{Oblique Lateral Wind Experiments}
\label{subsec:wind_exp}

A fan generates a 5\,\si{m/s} oblique wind (Fig.~\ref{fig:wind_exp}), loading both kinds of axes at once: an unmatched lateral force that pushes the vehicle off station, and matched aerodynamic moments on roll and pitch. By Proposition~\ref{prop:matched_unmatched}, lateral rejection is limited by the tilt path and tuned through $\lambda$. Table~\ref{tab:wind_controllers} lists the configurations and Table~\ref{tab:wind_perf} the results. 

\begin{table}[tbh]
\centering
\caption{\small{Controller assignments for oblique wind experiments}}
\setlength{\tabcolsep}{3pt}
\begin{tabular}{lccc}
\hline
\textbf{Controller config.} & \textbf{$x$-$y$ Pos.} & \textbf{$z$ Pos.} & \textbf{Attitude ($\phi,\theta,\psi$)} \\
\hline
PID-xyz/PID-att   & PID & PID & PID \\
SMC-xyz/PID-att   & SMC & SMC & PID \\
SMC-xy/PID-att    & SMC & PID & PID \\
SMC-xyz/SMC-att   & SMC & SMC & SMC ($\phi,\theta$), PID ($\psi$) \\
PID-xyz/SMC-att   & PID & PID & SMC ($\phi,\theta$), PID ($\psi$)  \\ 
\hline
\end{tabular}
\label{tab:wind_controllers}
\end{table}

\begin{figure}[tb]
    \centering
    \includegraphics[width=1\linewidth]{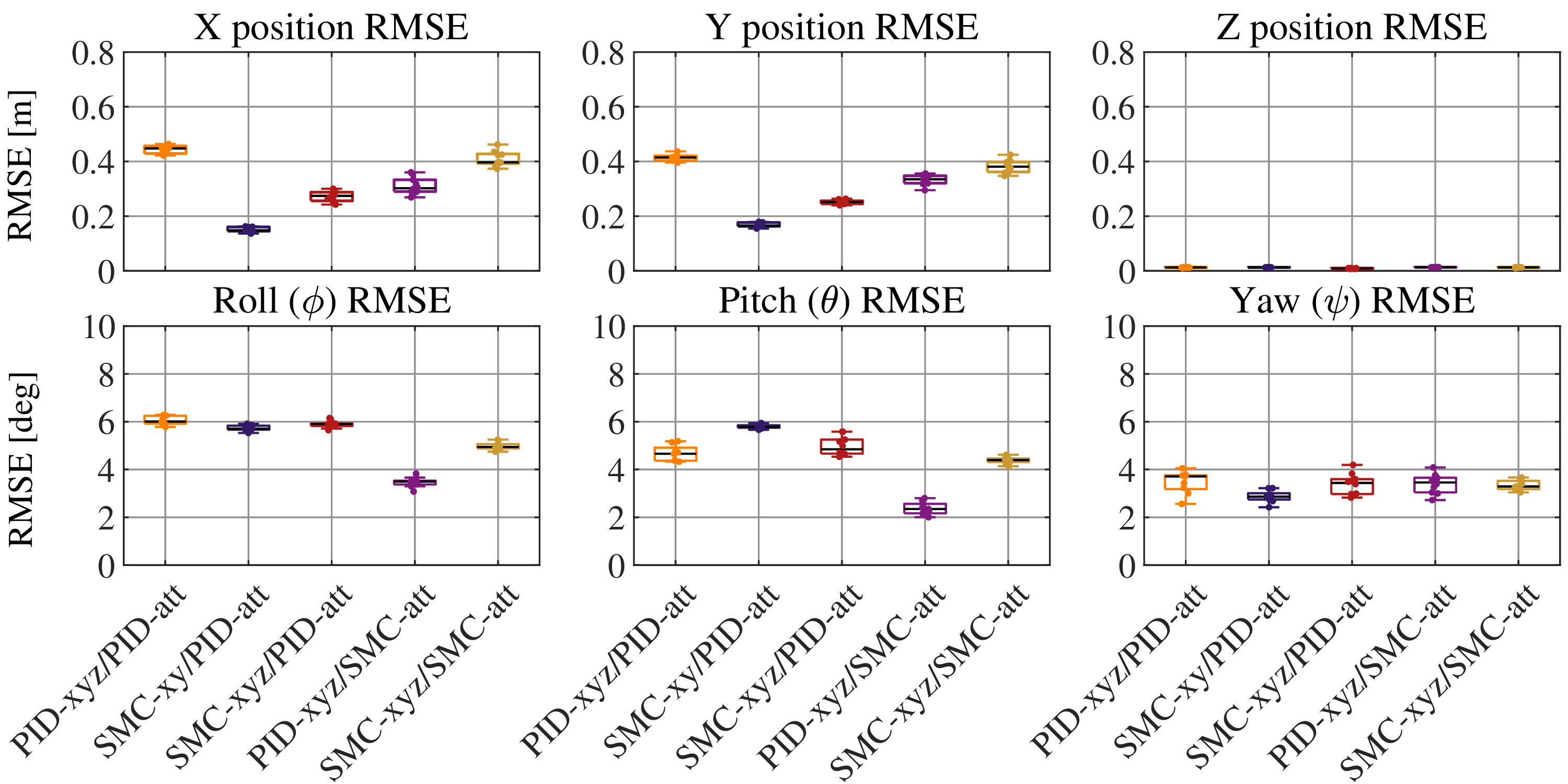}
    \caption{\footnotesize Per-axis RMSE for the oblique wind experiments with motion-capture feedback (Sec.~\ref{subsec:wind_exp}). Dots show individual trials, black lines medians, and bars the 25th--75th percentiles.
    \label{fig:wind_data_vicon_rmse}}
    \vspace{-2mm}
\end{figure}

\begin{table*}[hbtp]
\centering
\caption{\small{Tracking performance comparison (mean $\pm$ std.) for oblique wind disturbance experiments. Best-performing cases in bold.}}
\label{tab:wind_perf}
\begin{tabular}{lccccc}
    \hline
    \textbf{Metric} & \textbf{PID-xyz/PID-att} & \textbf{SMC-xy/PID-att} & \textbf{SMC-xyz/PID-att} & \textbf{PID-xyz/SMC-att} & \textbf{SMC-xyz/SMC-att} \\
    \hline
    \multicolumn{6}{l}{\textit{Wind-critical axes (lateral position \& attitude)}} \\
    $x$ RMSE (\si{m})    & $0.444 \pm 0.015$ & $\mathbf{0.150 \pm 0.009}$ & $0.272 \pm 0.018$ & $0.312 \pm 0.030$ & $0.410 \pm 0.028$ \\
    $y$ RMSE (\si{m})    & $0.413 \pm 0.012$ & $\mathbf{0.167 \pm 0.008}$ & $0.252 \pm 0.008$ & $0.333 \pm 0.018$ & $0.382 \pm 0.024$ \\
    Roll RMSE (deg) & $6.06 \pm 0.18$   & $5.72 \pm 0.13$ & $5.90 \pm 0.15$ & $\mathbf{3.47 \pm 0.20}$ & $4.98 \pm 0.17$ \\
    Pitch RMSE (deg)& $4.67 \pm 0.33$   & $5.79 \pm 0.08$ & $4.98 \pm 0.39$ & $\mathbf{2.38 \pm 0.25}$ & $4.39 \pm 0.15$ \\
    \hline
    \multicolumn{6}{l}{\textit{Less-affected axes (altitude and yaw)}} \\
    $z$ RMSE (\si{m})    & $0.0125 \pm 0.0012$ & $0.0129 \pm 0.0003$ & $\mathbf{0.0089 \pm 0.0013}$ & $0.0133 \pm 0.0004$ & $0.0125 \pm 0.0012$ \\
    Yaw RMSE (deg)  & $3.47 \pm 0.47$   & $\mathbf{2.87 \pm 0.25}$ & $3.38 \pm 0.44$ & $3.40 \pm 0.41$ & $3.34 \pm 0.21$ \\
    \hline
\end{tabular}
\vspace{-4mm}
\end{table*}

The configuration selected by our rule, SMC-xyz/PID-att, improves every excited position axis over the built-in PID, reducing $x$ and $y$ RMSE by 39\% and giving the lowest altitude error of all configurations (Table~\ref{tab:wind_perf}). Removing the switching term on $z$ (SMC-xy/PID-att) reduces lateral error further, to 66\% and 60\% below PID, at the cost of altitude: while the vehicle is tilted against the wind, thrust switching perturbs the lateral force, which the lateral loop must then reject. PID-xyz/SMC-att tracks the commanded tilt more accurately and gives the best roll and pitch, but the lateral error is set by the outer loop generating that command, which stays at PID. Uniform SMC performs worst among the SMC variants laterally, because switching at both position and attitude levels works against itself through the tilt coupling. Robustness must therefore be concentrated where the disturbance acts, not spread across every axis. Yaw is unexcited and left at PID, and shows little variation across configurations.

\subsection{Fully Onboard (Optical Flow State Estimation) Tracking Performance}
\label{subsec:optical_flow_exp}

The preceding experiments used motion capture for position feedback. To test the method under onboard state estimation, we repeat the wind experiment (Fig.~\ref{fig:wind_exp}) with position estimated fully onboard, fusing optical flow~\cite{Crazyflie} in the onboard EKF. Ground truth is streamed separately from Vicon Tracker 4.0~\cite{ViconUDP} over UDP. We compare the selective SMC variants against PID and, for this experiment, the Crazyflie's built-in INDI controller~\cite{SmeurINDIAttitude} with its default parameters and its own position loop. 

\begin{figure}[tbh]
    \centering
    \includegraphics[width=1\linewidth]{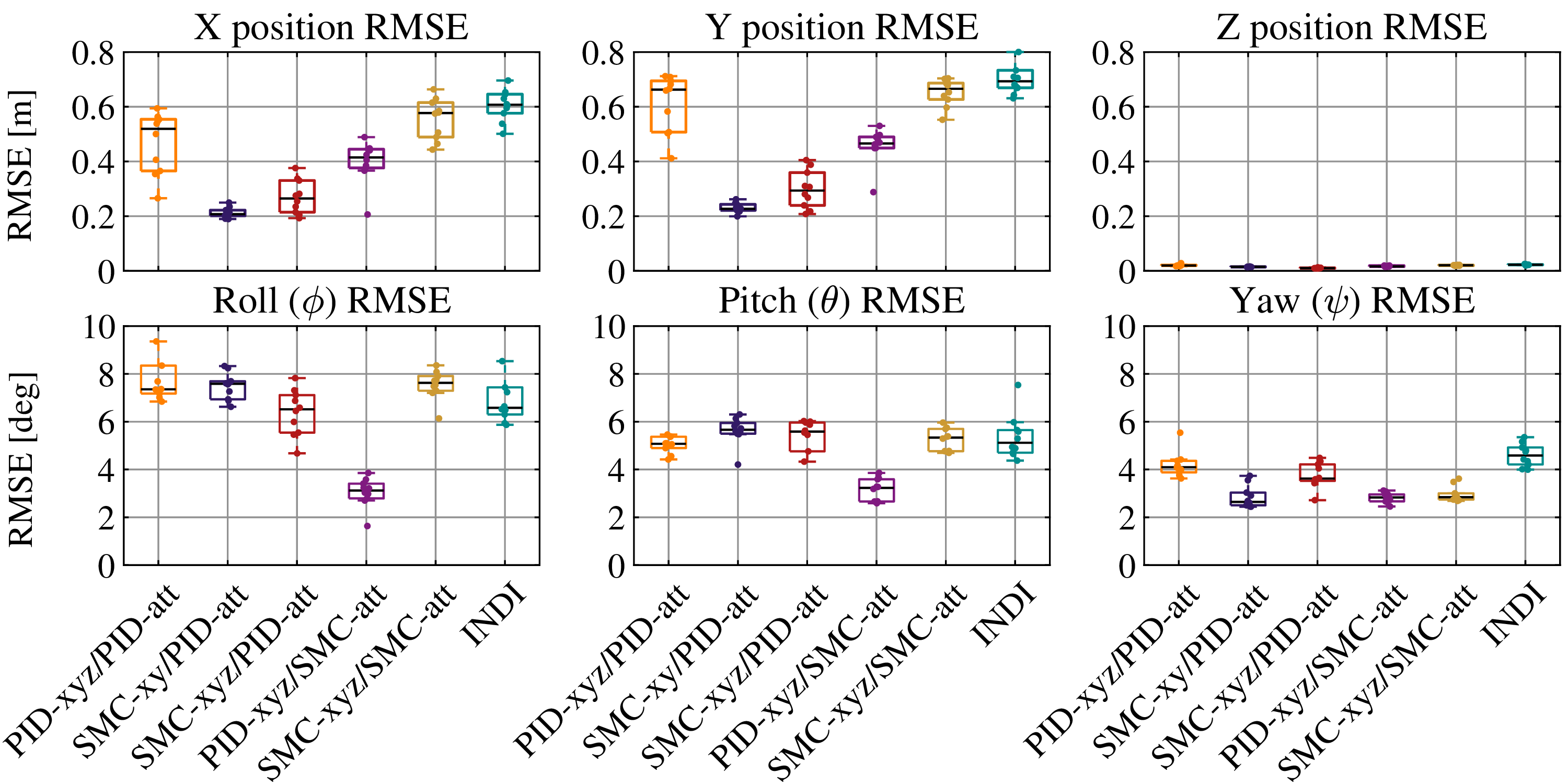}
    \caption{\footnotesize Per-axis RMSE for the oblique wind experiments with fully onboard optical-flow estimation (Sec.~\ref{subsec:optical_flow_exp}), including INDI. Dots show individual trials, black lines medians, and bars the 25th--75th percentiles.
    \label{fig:wind_data_optical_flow_rmse}}
    \vspace{-2mm}
\end{figure}

Moving from motion capture to optical flow raises the RMSE across all controller configurations, since onboard estimation is noisier than motion capture (cf. Figs.~\ref{fig:wind_data_vicon_rmse} and~\ref{fig:wind_data_optical_flow_rmse}). Their relative performance, however, is unchanged. Placing SMC on the lateral axes again tracks them best (Fig.~\ref{fig:wind_data_optical_flow_rmse}). 

On the attitude axes, PID-xyz/SMC-att gives the lowest RMSE, while the remaining configurations, including PID and INDI, perform comparably. Position is where the controllers separate most. INDI relies on gyroscope differentiation and division by the control effectiveness, both of which amplify the noise added once estimation moves onboard, so its position RMSE grows sharply while that of selective controllers stays substantially smaller. Our approach retains its benefit under the onboard sensing available on a deployed vehicle. This is consistent with the switching term treating the added estimation noise as part of the residual $\rho$ (Theorem~\ref{thm:bounded_dist}), the same way it treats the disturbance.

\subsection{Computational Cost}
\label{subsec:compute}

Table~\ref{tab:compute_cost} reports compute cost obtained by executing the compiled firmware controllers instruction by instruction in an emulator during simulated hover, with cycles estimated from Cortex-M4 instruction timings (168\,\si{MHz} core). The cascade runs its position loops on every fifth update, whereas INDI runs its full computation on every update, so we report both tick types and their average. Each switching term replaces a PID loop rather than adding to it, and the SMC update is lighter than the PID update, so every SMC configuration requires fewer cycles than PID on average. The savings grow with the number of loops replaced and how often they run. Replacing the 500\,\si{Hz} attitude loops lowers every update, whereas replacing the 100\,\si{Hz} position loops lowers only the full ticks. Selective SMC uses less of the processor than both PID and INDI under payload and a comparable share under wind, and all controllers use under 1\%. Our approach therefore adds robust rejection on the selected axes at no additional computational cost.

\begin{table}[tbh]
\centering
\caption{\small Cycles per control update (168\,\si{MHz} core). Full ticks run the position and attitude loops together at 100\,\si{Hz}, and attitude-only ticks fill the remaining 500\,\si{Hz} updates. The average weights one full tick per four attitude-only ticks, and CPU load is the processor share at the 1\,\si{kHz} stabilizer rate. Selective rows use the configurations chosen by Algorithm~\ref{alg:deploy} for each disturbance.}
\setlength{\tabcolsep}{3pt}
\begin{tabular}{lcccc}
\hline
\textbf{Controller} & \textbf{Full} & \textbf{Att.-only} & \textbf{Average} & \textbf{CPU load} \\
\hline
PID & 5989 & 2104 & 2881 & 0.88\% \\
Selective SMC (payload) & 5289 & 1666 & 2391 & 0.73\% \\
Selective SMC (wind) & 5207 & 2104 & 2725 & 0.83\% \\
Uniform SMC & 4769 & 1666 & 2287 & 0.70\% \\
INDI~\cite{SmeurINDIAttitude} & 2675 & 2671 & 2672 & 0.82\% \\
\hline
\end{tabular}
\label{tab:compute_cost}
\vspace{-4mm}
\end{table}

\section{Discussion and Conclusion}
We have shown that the benefit of robust control on an underactuated vehicle is confined to the axes that a disturbance excites and that the actuators can oppose. We formalize the classical matched and unmatched distinction as a per-axis deployment rule that identifies those axes from a small set of PID hover flights and determines which parameter to tune on each. Hardware experiments validated each of the rule's predictions. The switching gain reduced error on matched axes and had no effect on unmatched ones, where the surface slope alone set the floor. Under a slung payload and an oblique wind, tracking improved on exactly the axes the rule selected. When state estimation moved fully onboard, selective SMC held its advantage as INDI degraded, at no added computational cost over PID. 

In practice, this replaces trial-and-error tuning with a short procedure. Hover flights show which axes are excited, and the actuation geometry of each axis tells the practitioner whether to adjust the switching gain or the surface slope. If a lateral axis is deployed, the attitude loops underneath it stay on PID, because switching at both levels works against itself. None of this requires a disturbance model or new estimator. Where controllers differ is in how they perform under realistic sensing. With onboard estimation, the bound-based switching term proved more robust than an inversion that depends on differentiated gyro signals.

The approach has limits that point to future work. Our analysis and experiments focus on near-hover regulation, and identifying the excited axes requires flights under a representative instance of the disturbance. Running the excitation test online would remove the need for these dedicated flights, letting a vehicle enable or disable switching terms per axis as disturbances change mid-mission. Since the rule decides where robustness belongs rather than which law provides it, observer-based schemes such as DOBs and ESOs could be placed within the same framework, as could learned models absorbing residual dynamics on top of the SMC baseline. More broadly, the procedure applies to other underactuated systems with cascaded control, where actuation geometry determines which disturbances can be opposed directly.

\addtolength{\textheight}{-12cm}   




\bibliographystyle{IEEEtran}
\bibliography{references}

\end{document}